\documentclass[runningheads]{llncs}

\usepackage{array}
\usepackage{tabularx}
\usepackage{xcolor}

\usepackage{accv}

\usepackage{accvabbrv}

\usepackage{graphicx}
\usepackage{booktabs}

\usepackage[accsupp]{axessibility}  

\usepackage[pagebackref,breaklinks,colorlinks,citecolor=accvblue]{hyperref}

\usepackage{orcidlink}

\begin{document}
\sloppy

\title{A Proof-of-Concept Study of Weakly Supervised Labeling of Fine-Grained EEG Components for Artifact Attenuation} 

\titlerunning{Weakly Supervised Labeling of Fine-Grained EEG Components}

\author{Lu Wang-Nöth\inst{1}\orcidID{0009-0002-7443-121X} \and
Hai Huang\inst{1}\orcidID{0000-0001-8745-8142}\and \\
Philipp Heiler\inst{2} \and
Shuqiong Wu \inst{3}\orcidID{0000-0003-1501-9719} \and \\
Liyun Zhang \inst{4}\orcidID{0000-0003-3174-6588} \and
Helmut Mayer\inst{1}\orcidID{0000-0002-9439-2695}}

\authorrunning{L.~Wang-Nöth et al.}

\institute{Institute for Applied Computer Science, \\ University of the Bundeswehr Munich, Munich, Germany \email{lu.wang-noeth@protonmail.com} \and
brainboost GmbH, Munich, Germany \and 
Graduate School of Engineering Science, The University of Osaka, Osaka, Japan \and
Graduate School of Frontier Sciences, The University of Tokyo, Tokyo, Japan
}

\maketitle

\begin{abstract}
    Electroencephalography (EEG) is highly susceptible to electromyographic (EMG) artifacts, whose temporal heterogeneity and spatial-spectral overlap with neural activity can leave mixed sources after blind source separation. Existing artifact-removal methods are further limited by scarce reliable component-level ground truth: expert annotations are costly and subjective, while no established method provides realistic simulation-based ground truth for EMG contamination in multichannel scalp EEG.
    To address these limitations, we propose a framework combining a frequency-aware high-dimensional representation with Multi-Instance Learning. The representation unfolds separated components into frequency-resolved intra-components, creating a space in which mixed neural and muscular activity becomes more separable, while the weakly supervised learning formulation enables artifact-likelihood scores for individual intra-components to be learned from epoch-level labels without finer-grained ground truth.
    The resulting intra-component classifier supports fine-grained EMG artifact detection and score-guided attenuation. 
    Experiments on held-out subjects show that the framework learns informative intra-component scores and reduces artifact-related spectral deviations most clearly for jaw tension, with moderate effects for raising eyebrows and limited effects for frowning.

\end{abstract}

\section{Introduction}
\label{sec:intro}

    Electroencephalography (EEG) measures neural activity with high temporal resolution and is widely used in cognitive neuroscience, clinical research, and brain--computer interfaces (BCIs). However, EEG recordings are highly susceptible to non-neural artifact contamination. Among these artifacts, electromyographic (EMG) activity is particularly challenging because muscle activity is often broadband, transient, spatially heterogeneous, and partially overlapping with
    neural activity in both frequency and sensor space~\cite{goncharova2003emg,Muthukumaraswamy2013}. Consequently, EMG artifacts can distort spectral estimates and compromise downstream analyses, including neural decoding, biomarker discovery, and task-related spectral interpretation.
    
    Independent component analysis (ICA) is one of the most widely used approaches for EEG artifact correction ~\cite{makeig1995independent,delorme2004eeglab}.
    ICA decomposes multichannel EEG into statistically independent components, after which artifact-related components can be rejected manually or automatically. Several automated approaches have been proposed, including rule-based and learned classifiers for independent components~\cite{winklerAutomaticClassificationArtifactual2011,pion2019iclabel}. Despite these advances, ICA-based cleaning remains limited when neural and muscular activity are not cleanly separated into distinct components, particularly in low-density EEG settings ($<$32 channels)~\cite{bono2016hybrid,lopezHAPPILEEHAPPELow2022}. In such cases, rejecting an entire component may remove useful neural information, whereas retaining it may leave residual EMG contamination.
    Beyond ICA, EEG artifact correction has also been addressed using regression, wavelet denoising, subspace reconstruction, and hybrid approaches~\cite{gratton1983regression,castellanos2006wica,mullen2015asr,uriguenEEGArtifactRemoval2015,Jiang2019}.
    Regression-based methods can be effective when appropriate reference channels are available, but they depend on assumptions about the relationship between reference channels and contaminated EEG. Wavelet-based and ICA--wavelet methods can attenuate transient artifacts, but their performance depends on the choice of wavelet basis, thresholding strategy, and component-selection criteria. Artifact subspace reconstruction and related subspace methods can suppress large-amplitude nonstationary artifacts, but their behavior depends on calibration data and rejection thresholds. These methods are valuable for EEG preprocessing, yet they typically make correction decisions at the level of channels, time windows, or entire components. These operating scales can be too coarse when neural and muscular activity coexist within the same unit. This motivates artifact correction at a finer level.
    
    A second major challenge is the lack of reliable component-level ground truth. Prior work has often relied on expert annotations, which require substantial manual effort and introduce subjective bias. Moreover, recent studies suggest that artifact correction and rejection do not always improve downstream EEG analyses and may even reduce decoding performance when removed artifacts covary with the experimental condition or when neural signal is removed together with artifacts ~\cite{delormeEEGBetterLeft2023,kessler2025preprocessing}. Synthetic data could provide theoretical ground truth, but it may not capture the spatial, spectral, and anatomical variability present in realistic recording conditions~\cite{uriguenEEGArtifactRemoval2015,Jiang2019}. This limitation is especially relevant for EMG contamination, whose spatial distribution across electrodes depends on muscle source, head anatomy, electrode montage, and task context. Moreover, to our knowledge, no established method currently provides validated, realistic simulations of how EMG artifacts appear in multi-channel EEG. Such a simulation would prescribe source structures and spatial distributions which should be inferred from real recordings.
    
    To address these limitations, we combine the frequency-aware high-dimensional representation proposed in~\cite{wang2025beyond} with weakly supervised Multi-Instance Learning (MIL)~\cite{dietterich1997mil}. MIL is a weakly supervised framework in which labels are provided for bags of instances rather than for individual instances. In the classical binary formulation, a positive bag contains at least one positive instance, whereas a negative bag contains none, enabling instance-level scores to be learned from bag-level supervision.    
    Instead of treating each source-separated component as an indivisible unit, the high-dimensional representation unfolds each component along the frequency dimension into frequency-resolved intra-components. This design is motivated by the observation that neural and muscular activity that remains mixed at the component level may become more separable when represented at the intra-component level. Whereas~\cite{wang2025beyond} introduced the frequency-aware representation, the present work uses this representation as the instance space for weakly supervised EMG artifact learning and evaluates whether the resulting intra-component scores can guide fine-grained artifact attenuation.
    We formulate artifact detection in this representation as a weakly supervised MIL problem. Each EEG epoch, defined as a short EEG segment, is treated as a bag, and the frequency-resolved intra-components derived from that epoch are treated as instances. The model is trained using only epoch-level labels: resting eyes-open (EO) and eyes-closed (EC) epochs are treated as putatively clean examples, whereas epochs containing instructed jaw tension, raising eyebrows, or frowning are treated as artifact-contaminated examples. Because these weak labels are defined by experimental condition, the model could in principle learn condition-specific neural or behavioral differences rather than EMG contamination alone. We reduce potential subject-specific confounding through subject-wise splits and address this limitation in the interpretation of the learned scores, using reconstruction-based spectral validation rather than treating them as definitive artifact labels. By learning to predict epoch-level contamination from bags of intra-components, the shared encoder and score head can subsequently be reused to assign scores to individual intra-components without requiring component-level or intra-component-level annotations. Under this weak-supervision formulation, the model is encouraged to assign high scores to intra-components that are predictive of artifact-contaminated epochs. These scores should be interpreted as artifact-likelihood scores rather than verified annotations. 
    
    Rather than as a benchmark against conventional epoch-level classifiers, epoch-level performance serves primarily to support the central objective of this work: learning interpretable intra-component artifact-likelihood scores from weak supervision for fine-grained artifact removal.
    Previous studies for artifact removal differ concerning datasets, channel montages, artifact definitions, and reference. Available benchmarks address narrower settings, including artifact-segment detection~\cite{9353647}, component classification~\cite{pion2019iclabel}, and single-channel semi-synthetic denoising~\cite{zhangEEGdenoiseNetBenchmarkDataset2021}. To our knowledge, none pairs realistic multi-channel EEG–EMG mixtures with clean references under a matching protocol. Therefore, the present study does not demonstrate superiority over existing artifact-removal methods; rather, it should be viewed as a proof of concept for weakly supervised intra-component artifact-likelihood learning.
    
    The main contributions of this work are as follows:
    \begin{itemize}
    
        \item We formulate EMG artifact detection as a weakly supervised MIL problem in which EEG epochs serve as bags and frequency-resolved intra-components serve as instances.
    
        \item We show that epoch-level supervision can induce intra-component artifact-likelihood scores without relying on costly and subjective fine-grained annotations or currently unavailable realistic simulation-based ground truth.
            
        \item We evaluate whether score-guided attenuation of high-scoring intra-components reduces artifact-related spectral deviations, with the clearest effects observed for jaw tension, moderate effects for raising eyebrows and limited effects for frowning.
    \end{itemize}
      
\section{Data and Methods}

    \subsection{Data Collection and Preprocessing}
        
        To evaluate the proposed framework, we analyzed EEG epochs containing representative physiological artifacts. Following~\cite{wang2025much}, we focused on three EMG-related artifact types: jaw tension, raising eyebrows, and frowning. EO and EC resting-state epochs were used as clean EEG classes. Combining newly collected data with publicly available datasets~\cite{wang2025much,Wang-NoethEtAll2024BREAD}, the final dataset included recordings from 122 participants.
        All participants were informed about the purpose and procedures of the study and provided written informed consent. The study was approved by the relevant ethics committee and conducted in accordance with the principles of the Declaration of Helsinki and applicable local regulations.
        
        Each subject contributed three 5-second EO epochs, three 5-second EC epochs, and at least one epoch for each of the three artifact classes. Artifact epochs were designed to last 5 seconds; epochs shorter than 3 s were excluded from the analysis. The training set comprised 98 participants and the held-out test set comprised 24 participants, with no participant appearing in both sets. In total, 883 epochs were used for training, and 216 epochs were reserved as a held-out test set, which contained recordings from subjects who were not included in the training set. The present evaluation therefore assesses performance on held-out subjects within the available datasets; broader transferability across datasets, artifact types, mental states, and hardware configurations will be examined in future work. The training set consisted of 294 EO, 294 EC, 99 jaw-tension, 98 raising-eyebrows, and 98 frowning epochs. The test set consisted of 72 EO, 72 EC, 24 jaw-tension, 24 raising-eyebrows, and 24 frowning epochs.
        All epochs were recorded using a 25-channel international 10--10 EEG electrode montage, with Cz as the online reference. The recordings were preprocessed as follows:
        \begin{itemize}
            \item High-pass filtering at 1 Hz to remove low-frequency drifts;
            \item Notch filtering at 50 Hz and its harmonics up to 1001 Hz to attenuate power-line noise;
            \item Re-referencing to the common average across EEG channels.
        \end{itemize}

    \subsection{Intra-Components and Feature Extraction}
        \begin{figure}[!htb]
            \centering
            \includegraphics[width=\linewidth, trim=0 40mm 0 0mm, clip]{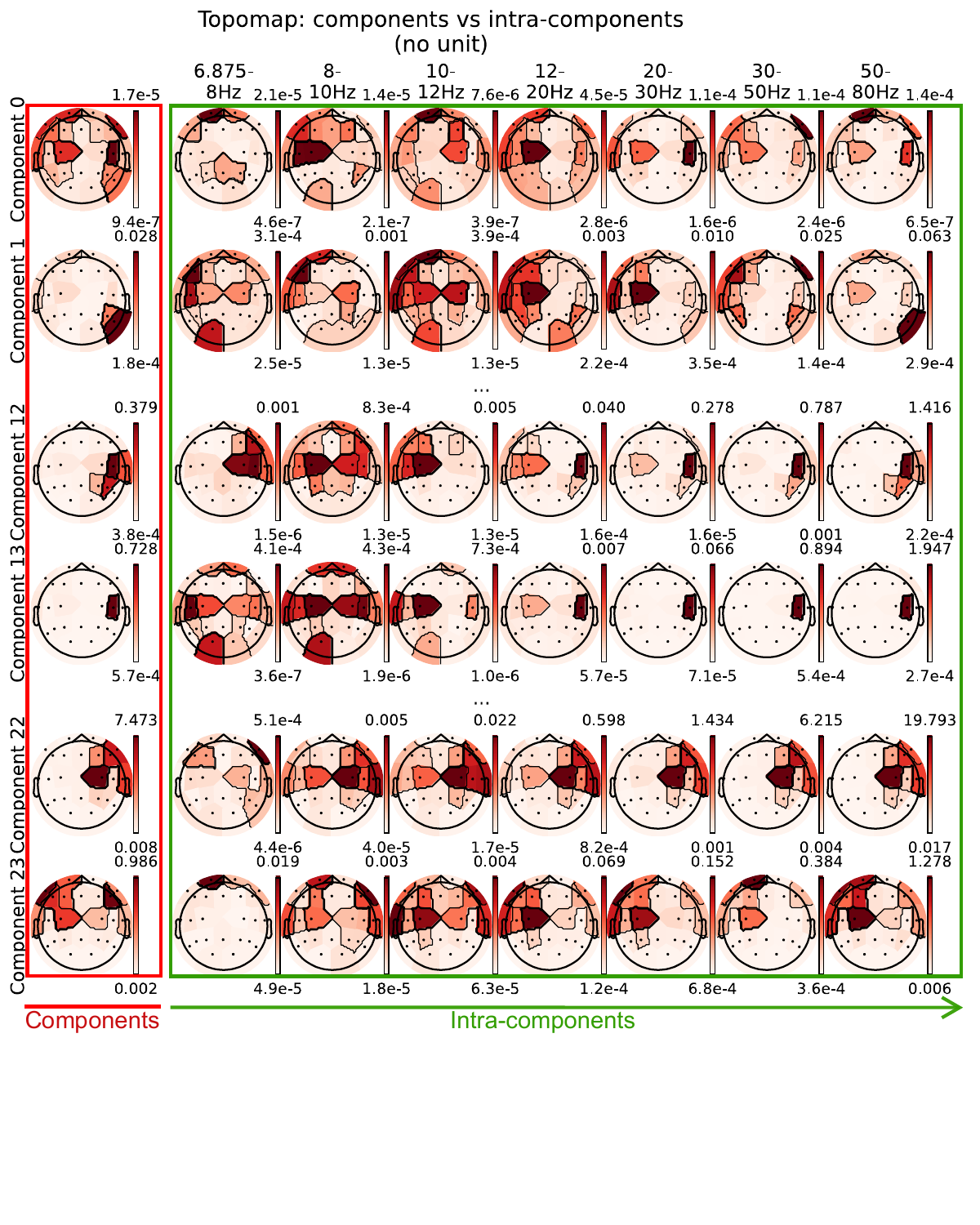}
            \caption{Unfolding \textcolor{red}{components} to \textcolor{OliveGreen}{intra-components} with components 0, 1, 12, 13, 22 and 23 as examples. For visual clarity, frequency bins are aggregated into standard EEG frequency bands: theta (6.875–8 Hz; corresponding to the available portion of the conventional 4–8 Hz theta band), low-alpha (8--10 Hz), high-alpha (10--12 Hz), low-beta (12--20 Hz), high-beta (20--30 Hz), low-gamma1 (30--50 Hz), and low-gamma2 (50--80 Hz). Following~\cite{wang2025beyond}, the ILRMA decomposition starts at 6.875 Hz to avoid linear dependencies across frequency; each topomap uses an individual color scale to show its spatial distribution; the numbers beside each topomap are the limits of its color range for unitless, rescaled projected power; hence, colors are not quantitatively comparable across maps. The intra-component representation enables frequency-wise separation of mixed neural and non-neural activity within components.}
            \label{fig:topomaps}
        \end{figure}

        {\bf Intra-Components.} 
        Intra-components were constructed using the procedure described by~\cite{wang2025beyond}. 
        Each epoch was first standardized by subtracting its mean and dividing by its standard deviation. A continuous wavelet transform was then computed for all channels using a Generalized Morse wavelet with parameters $\gamma = 3$ and $\beta = 5$, logarithmic maximal scales, and 16 voices per octave.
        The complex wavelet coefficients were arranged as $time \times frequency \times channel$ and downsampled from 2048 Hz to 512 Hz before source separation for more efficient computation. A subset of 95 wavelet frequency bins was retained for independent low-rank matrix analysis (ILRMA), corresponding approximately to 6.9--403.5 Hz. However, only frequencies below 80 Hz, specifically 6.875--77.75 Hz, were analyzed in the final results, focusing on the main spectral range relevant to EEG activity. ILRMA was then applied to the selected time-frequency representation using 24 components, matching the number of EEG channels after excluding the reference channel. ILRMA was run for 40 iterations with 40 low-rank bases and without projection-back scaling.
        For each epoch, the procedure produced complex-valued component activations $Y_{fst}$ and corresponding mixing matrices $A_{fcs}$, where $f, s, t$ and $c$ denote frequency, component, time and channel, respectively.
        The two three-dimensional outputs were combined through multiplication according to \cref{eq_dimtrans_1} to obtain the four-dimensional tensor $Y^*_{fstc}$~\cite{wang2025beyond}, thereby preserving information from all four dimensions. The components were then unfolded along the frequency dimension, such that each frequency--component pair $(f,s)$ defined one intra-component with dimensions time $\times$ channel.
        \begin{align}
            A_{fcs} Y_{fst}  &= Y^*_{fstc} \label{eq_dimtrans_1}.
        \end{align} 

        For each intra-component, two complementary feature representations were extracted: a spatial topographic map (topomap) based on channel dimension and a handcrafted temporal representation based on the time dimension. Each epoch yields $frequency\times components$ intra-components.
        
        {\bf Spatial Features.}
        Spatial features were computed separately for each intra-component within each epoch. Its complex-valued mixing coefficients across EEG channels were represented by their normalized magnitude, cosine of phase, and sine of phase. Magnitudes were normalized by their 95th percentile across channels within that intra-component and clipped to $[0,1]$. Each representation was then spatially interpolated to form one channel of a three-channel topomap. These channel-level values were interpolated onto a $32\times 32$ scalp grid using the MNE-Python~\cite{GramfortEtAl2013a} standard 10-20 montage, cubic interpolation, and masking outside the head outline.
        
        {\bf Temporal Features.} 
        Temporal features were computed from the corresponding complex source time series. For each intra-component, the log-magnitude time course was resampled to 256 samples and standardized. Magnitude-based descriptors included mean activity, fraction of samples exceeding a burst threshold, longest burst duration, concentration of energy in the top 10\% of samples, overall temporal variation, autocorrelation at relative lags of 0.02, 0.05, 0.10, and 0.20, and windowed summaries over 4, 8, and 16 temporal windows. Phase-based descriptors were extracted from the instantaneous phase and phase increments, including circular concentration, phase increment variability, mean absolute phase increment, amplitude-weighted phase concentration, phase similarity at the same relative lags, and windowed phase summaries.
        In total, 66 handcrafted temporal features were extracted for each intra-component: 45 magnitude-based features and 21 phase-based features. Details are provided in Appendix~\ref{Appendix_temporal_features}. 

    \subsection{Training Architecture}
    
        The classifier was implemented as a weakly supervised MIL model. Each EEG epoch was treated as a bag of intra-components, where each intra-component corresponds to one frequency--component pair. For epoch $e$, the bag is denoted as $\mathcal{B}^{(e)} = \{x_i^{(e)}\}_{i=1}^{N}$, where $N = F \times S$, with $F$ frequency bins and $S$ components.
        Each intra-component was represented by three inputs: a 3-channel spatial topomap $T_i \in \mathbb{R}^{3 \times 32 \times 32}$, a 66-dimensional handcrafted temporal feature vector $v_i \in \mathbb{R}^{66}$, and a normalized scalar frequency coordinate $f_i \in \mathbb{R}$. The topomap and temporal features were encoded by a convolutional branch and a multilayer perceptron, respectively. Their outputs were concatenated with $f_i$ and projected to a 64-dimensional intra-component embedding $h_i \in \mathbb{R}^{64}$. 
        A lightweight frequency-context encoder, consisting of two 1D convolutional layers along the frequency axis for each source, then modeled local continuity across neighboring frequency bins before the sigmoid score head produced an intra-component artifact-likelihood score $p_i \in [0,1]$.
        
        Training used only weak epoch-level labels: EO and EC epochs were labeled as putatively clean, $y^{(e)}=0$, whereas jaw tension, raising eyebrows, and frowning epochs were labeled as artifact-contaminated, $y^{(e)}=1$. Intra-component scores were aggregated by top-$k$ MIL pooling. We set $k=2\%$ as a heuristic intended to preserve sparse artifact evidence for weak contamination; the highest-scoring 2\% of intra-components were averaged to obtain the epoch-level contamination score $\hat{y}^{(e)}$. 
        Training used soft bag targets $\tilde{y}^{(e)}=0.05$ for clean-labeled epochs and $\tilde{y}^{(e)}=0.90$ for contaminated-labeled epochs; binary cross-entropy (BCE) was computed between $\hat{y}^{(e)}$ and $\tilde{y}^{(e)}$.
        The full objective additionally encouraged sparse high-score assignments, separation between clean and contaminated epoch scores, and smoothness across adjacent frequency bins. Specifically, it combined the BCE loss with sparsity, ranking, and frequency-smoothness regularization terms. The sparsity term penalized high average intra-component scores, the ranking term encouraged contaminated epochs to receive higher MIL scores than clean epochs, and the frequency-smoothness term penalized abrupt score changes across adjacent frequency bins for the same source. Training used AdamW optimization, balanced sampling, a batch size of 4, 30 complete passes over the training set, and a frequency-smoothness weight of $\lambda_{\mathrm{freq}}=0.10$.
        After training, the MIL pooling layer was discarded. The shared encoder and sigmoid score head were reused for intra-component classification, assigning each intra-component an artifact-likelihood score $p_i$. Although the framework permits threshold selection on validation data, the proof-of-concept cleaning evaluation used a fixed threshold of $\tau=0.5$, the sigmoid midpoint, to define a binary attenuation mask. Intra-components with $p_i\ge\tau$ were treated as artifact-like and attenuated during reconstruction.
        Details of the concrete architecture, including the number of layers and hidden dimensions, are provided in Appendix~\ref{Appendix_architecture}.

        \begin{figure}[htb!]
            \centering
            \includegraphics[width=\linewidth, , trim=0 39mm 0 0mm, clip]{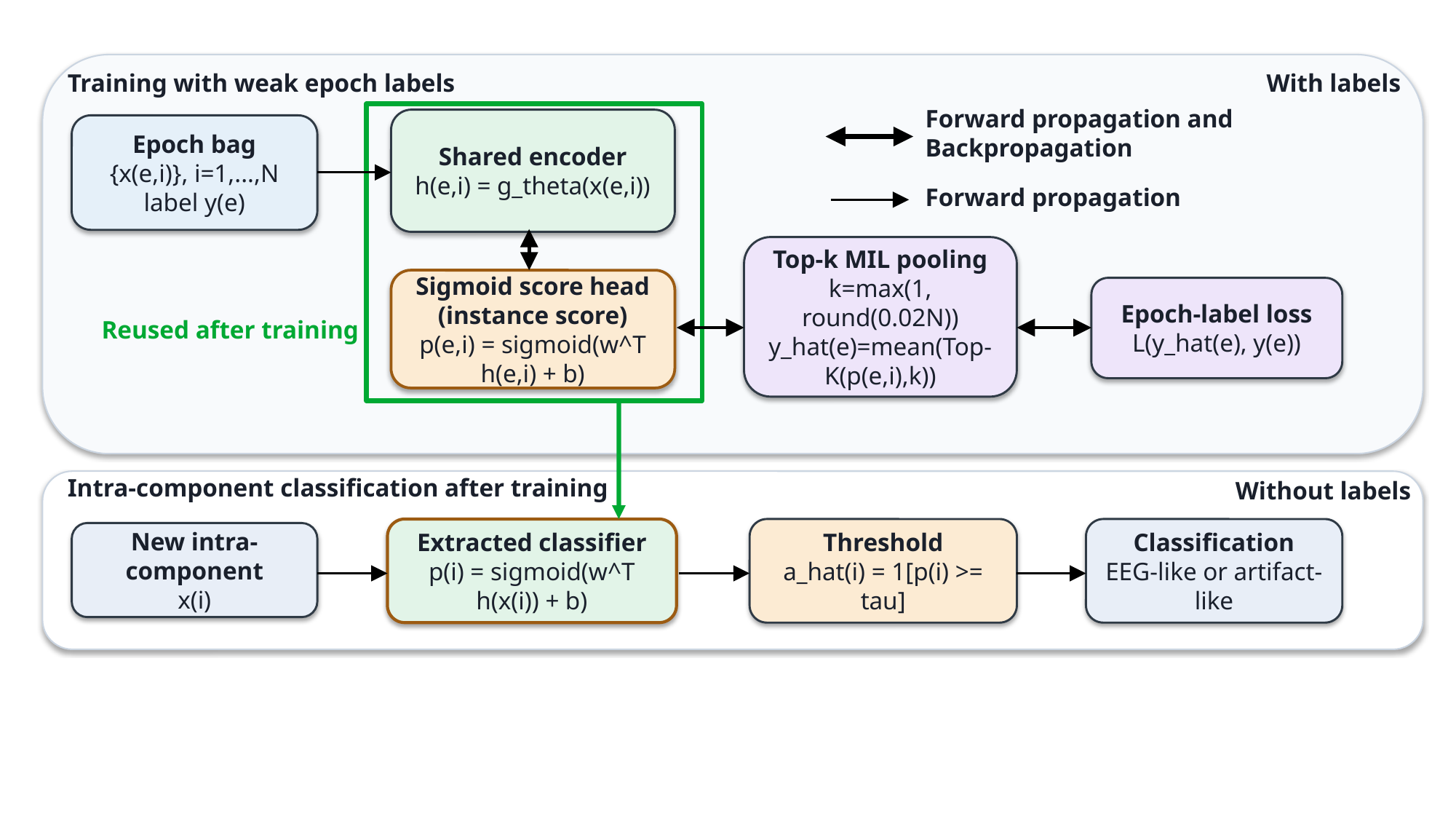}
            \caption{Weakly supervised MIL-based intra-component classifier. During training, each EEG epoch is treated as a bag of intra-components. A shared encoder maps each intra-component to an embedding, and a sigmoid head assigns an instance-level artifact-likelihood score. The highest-scoring 2\% of intra-components are averaged by top-$k$ MIL pooling to obtain the epoch-level contamination score, which is optimized against the weak epoch label. After training, the MIL pooling layer is discarded, and the encoder and score head are reused for intra-component classification. The framework permits threshold selection on validation data, while the proof-of-concept cleaning evaluation used a fixed threshold of $\tau=0.5$ to define a binary attenuation mask.}
            
            \label{fig:arch}
        \end{figure}

    \subsection{Score-Guided Removal and Reconstruction}
        The binary removal mask, denoted by $\hat{a}_{f,s}$, equals one when $p_{f,s}\geq\tau$ and zero otherwise. Artifact-like intra-components are removed by setting their source activations to zero:
        \begin{equation}
        \tilde{Y}_{f,s,t}
        =
        Y_{f,s,t}
        \left(1-\hat{a}_{f,s}\right).
        \end{equation}
        Thus, only intra-components below the artifact-likelihood threshold were retained for reconstruction.
        The cleaned channel-level spectrogram was then reconstructed from the retained
        intra-components using the ILRMA mixing matrix:
        \begin{equation}
        \tilde{X}_{f,t,c}
        =
        \sum_{s=1}^{S}
        A_{f,c,s}\tilde{Y}_{f,s,t},
        \end{equation}
        where \(c\) denotes the EEG channel.
        Therefore, the cleaning operation differs from the full reconstruction only by the
        binary intra-component mask applied to \(Y_{f,s,t}\) before summation.
        This procedure yields a cleaned complex spectrogram
        \(\tilde{X}_{f,t,c}\) in the same frequency--time--channel representation as
        the original epoch spectrogram.

    \subsection{Metrics} \label{sec:metrics}

        We report both epoch-level classification and indirect reconstruction-based evidence for intra-component score-guided attenuation. Epoch-level classification provides the only directly supervised evaluation and serves as indirect support for the learned intra-component scores; the effect of intra-component attenuation is assessed indirectly through reconstruction-based spectral metrics.

        {\bf Standard Binary Classification Metrics.}    
        Standard binary classification metrics were reported for epoch-level classification.
        Epoch-level performance was evaluated from the MIL-pooled epoch prediction
        \(\hat{y}^{(e)} \in [0,1]\), which represents the predicted probability that
        epoch \(e\) is artifact-contaminated. The corresponding binary
        epoch label was denoted by \(y^{(e)} \in \{0,1\}\), where \(0\) indicates a
        clean epoch and \(1\) indicates an artifact-contaminated epoch.
        For threshold-based metrics, the predicted probability ($\hat{y}^{(e)}$) was converted into a binary prediction using a decision threshold ($\tau$). We reported BCE, balanced accuracy, recall, specificity, precision, and F1 score. BCE was used to assess the agreement between predicted epoch-level probabilities and labels. Recall measured the proportion of artifact-contaminated epochs correctly detected, whereas specificity measured the proportion of clean epochs correctly identified. Balanced accuracy was used to account for class imbalance by averaging recall and specificity. Precision quantified the proportion of predicted artifact-contaminated epochs that were truly contaminated, and the F1 score summarized the trade-off between precision and recall.
        In addition, threshold-independent performance was quantified using the receiver operating characteristic area under the curve (ROC AUC) and average precision (AP). ROC AUC measures the ability of the model to rank contaminated epochs above clean epochs across all possible thresholds, namely all unique predicted artifact scores in the held-out test set. AP summarizes the precision--recall curve and is particularly relevant when class frequencies are imbalanced.
           
        {\bf Time-Averaged Power-Spectrum Improvement.}    
        Because ground-truth intra-component labels were unavailable, cleaning performance was evaluated by comparing the time-averaged power spectra before and after removing intra-components classified as artifact-like. For each artifact epoch, the original artifact spectrogram \(X^{\mathrm{before}}\), cleaned spectrogram \(X^{\mathrm{after}}\), and same-subject EO reference \(X^{\mathrm{ref}}\) were compared after scale-correcting amplitudes to the reference. Note that the EO reference serves as a practical comparison target, rather than a true ground truth. Differences between the cleaned spectrum and the clean reference may reflect both residual artifact and genuine differences in brain state.
        For \(q \in \{\mathrm{before},\mathrm{after},\mathrm{ref}\}\), time-averaged power was computed as
        \begin{equation}
        P^{q}_{f,c} = \frac{1}{T}\sum_{t=1}^{T}\left|X^{q}_{f,t,c}\right|^2 .
        \end{equation}
        Below 80 Hz, the reference area and deviations from the reference were
        \begin{equation}
        A^{\mathrm{ref}}_{c}
        =
        \int_{6.875}^{77.75}\left|P^{\mathrm{ref}}_{f,c}\right|\,df,
        \quad
        D^{q}_{c}
        =
        \int_{6.875}^{77.75}
        \left|P^{q}_{f,c}-P^{\mathrm{ref}}_{f,c}\right|\,df,
        \quad
        q \in \{\mathrm{before},\mathrm{after}\}.
        \end{equation}
        Relative deviations were then defined as
        \begin{equation}
        d^{q}_{c}
        =
        \frac{D^{q}_{c}}{A^{\mathrm{ref}}_{c}+\epsilon},
        \quad
        q \in \{\mathrm{before},\mathrm{after}\},
        \quad
        \epsilon=10^{-12},
        \end{equation}
        and the main intra-component-level metric was the relative area improvement,
        \begin{equation}
        \Delta d_{c}
        =
        d^{\mathrm{before}}_{c}
        -
        d^{\mathrm{after}}_{c}.
        \end{equation}
        Positive values indicate that the cleaned spectrum became closer to the reference, whereas negative values indicate reduced similarity. The relative area improvement \(\Delta d_c\) was summarized by artifact class and EEG channel; class--channel medians are reported in \cref{sec:results}, with detailed descriptive statistics in Appendix~\ref{Appendix_stats}.
        
\section{Results} \label{sec:results}
    \subsection{Epoch-Level Results}
        These epoch-level results assess whether the weak labels provide a learnable bag-level supervision signal, which serves as indirect support for interpreting the learned intra-component scores.
        \Cref{tab:mil_test_results_threshold_05} summarizes the epoch-level test performance of the MIL model using a fixed decision threshold of 0.5, without systematic threshold optimization. The model obtained a balanced accuracy of 0.91, indicating that the weakly labeled clean and artifact-contaminated epochs were separable in the held-out-subject test set. Recall was 0.89 and specificity was 0.93, showing that the model detected most contaminated epochs while also correctly identifying most clean epochs. Precision reached 0.86, indicating that most epochs predicted as artifact-contaminated were true positives. The resulting F1 score of 0.88 demonstrates a balanced trade-off between precision and recall. The BCE value of 0.38 provides an additional measure of agreement between the predicted epoch-level probabilities and the labels.
        \Cref{tab:mil_test_results_threshold_range} summarizes the epoch-level test performance across different thresholds -- artifact scores observed in the test set, from 0.02 to 0.95. The ROC AUC and AP are 0.98 and 0.96, respectively, further indicating that the model could rank instructed artifact-contaminated epochs above putatively clean epochs in the held-out-subject test set.
        
        \begin{table}[t]
            \caption{Epoch-level test classification results for the MIL model at threshold 0.5}
            \label{tab:mil_test_results_threshold_05}
            \centering
            \begin{tabular}{|l|c|}
            \hline
            Metrics & Test @$\tau=0.5$ \\
            \hline\hline
            BCE & 0.38 \\
            Balanced accuracy & 0.91 \\
            Recall & 0.89 \\
            Specificity & 0.93 \\
            Precision & 0.86 \\
            F1 score & 0.88 \\
            \hline
            \end{tabular}
        \end{table}

        \begin{table}[t]
            \caption{Epoch-level test classification results for the MIL model across different thresholds}
            \label{tab:mil_test_results_threshold_range}
            \centering
            \begin{tabular}{|l|c|}
            \hline
            Metrics & Test @$\tau \in [0.02,0.95]$ \\
            \hline\hline
            ROC AUC & 0.98 \\
            Average Precision & 0.96\\
            \hline
            \end{tabular}
        \end{table}

    \subsection{Intra-Component-Level Results}

        \begin{table}[t]
        \centering
        \scriptsize 
        \setlength{\tabcolsep}{0pt}
        \renewcommand{\arraystretch}{0.92}
        
        \newcolumntype{A}{>{\centering\arraybackslash}p{0.09\textwidth}}
        \newcolumntype{C}{>{\centering\arraybackslash}p{0.07\textwidth}}

        \begin{tabular}{|A|C|C|C|C|C|C|C|C|C|C|C|C|}
        \hline
        Artifact & Fp1 & Fp2 & F9 & F7 & F3 & Fz & F4 & F8 & F10 & T9 & T7 & C3 \\
        \hline\hline
        JT & 5.09 & 7.38 & \textcolor{teal}{84.77} & \textcolor{teal}{111.53} & \textcolor{teal}{86.69} & \textcolor{teal}{25.21} & \textcolor{teal}{120.05} & \textcolor{teal}{144.94} & \textcolor{teal}{141.23} & \textcolor{teal}{43.53} & \textcolor{teal}{18.54} & \textcolor{teal}{95.97} \\
        RE & \textcolor{red}{-0.47} & \textcolor{teal}{17.97} & 0.30 & 2.27 & \textcolor{teal}{9.86} & 3.03 & \textcolor{teal}{10.03} & 3.33 & 0.66 & 0.31 & 0.52 & 1.89 \\
        Fr & \textcolor{red}{-0.71} & 0.24 & 0.02 & 0.02 & 0.06 & 0.02 & 0.02 & 0.04 & 0.03 & 0.00 & 0.00 & 0.02 \\
        \hline
        \end{tabular}
        
        \vspace{0.6em}
        
        \begin{tabular}{|A|C|C|C|C|C|C|C|C|C|C|C|C|}
        \hline
        Artifact & C4 & T8 & T10 & P9 & P7 & P3 & Pz & P4 & P8 & P10 & O1 & O2 \\
        \hline\hline
        JT & \textcolor{teal}{124.30} & \textcolor{teal}{14.68} & \textcolor{teal}{42.83} & 1.52 & 2.53 & \textcolor{teal}{11.75} & \textcolor{teal}{11.10} & \textcolor{teal}{16.89} & 4.22 & 0.70 & 2.47 & 1.45 \\
        RE & 1.72 & 1.06 & 0.31 & 0.32 & 0.55 & 0.70 & 0.83 & 0.76 & 0.90 & 0.46 & 0.67 & 0.54 \\
        Fr & 0.01 & 0.00 & 0.01 & 0.00 & 0.02 & 0.00 & 0.02 & 0.02 & 0.00 & 0.00 & 0.00 & -0.00 \\
        \hline
        \end{tabular}
        
        \caption{
        Channel-wise median relative area-difference improvement after removing artifact-like intra-components, grouped by artifact type: jaw tension (JT), raising eyebrows (RE), and frowning (Fr), with $N=24$ epochs per type. Values are reported in the units of the relative-deviation metric \(\Delta d_c\) and should not be interpreted as percentage improvement. Positive values indicate reduced spectral deviation from the same-subject EO reference, whereas negative values indicate increased deviation after cleaning. Green highlights mark pronounced positive improvements, and red marks undesirable changes. Jaw tension showed the strongest improvements over frontal, fronto-temporal, and central channels; raising eyebrows showed smaller mainly frontal improvements; frowning produced near-zero changes across channels.
        }
        \label{tab:area_improvement_by_channel}
        \end{table} 

        In this section, we report intra-component-level attenuation results using the fixed threshold 0.5. Because no component- or intra-component-level ground truth was available, the effect of attenuation was assessed using the time-averaged power-spectrum improvement metric described in \cref{sec:metrics}. Across the three artifact types, the clearest improvement was observed for jaw tension, with the strongest reductions over channels consistent with expected jaw-related muscle activity.
        \Cref{tab:area_improvement_by_channel} summarizes the relative spectral area improvement, \(\Delta d_c\), computed against the clean reference spectrum for each channel. Positive values indicate reduced deviation from the reference after intra-component removal, whereas negative values indicate increased deviation. Jaw tension showed the largest improvements, especially over frontal, fronto-temporal, and central channels, with the five highest values at F8 (144.94), F10 (141.23), C4 (124.30), F4 (120.05), and F7 (111.53). This pattern is consistent with the expected frontal and temporalis involvement of jaw-related muscle activity. 
        Raising eyebrows showed smaller effects, with the strongest improvements at Fp2 (17.97), F4 (10.03), and F3 (9.86), whereas Fp1 showed a slight negative value close to zero. Frowning produced near-zero changes across channels, indicating limited spectral improvement for this artifact class.

        \begin{figure}[t!]
        \includegraphics[width=\linewidth, trim=0 20mm 0 0mm, clip ]{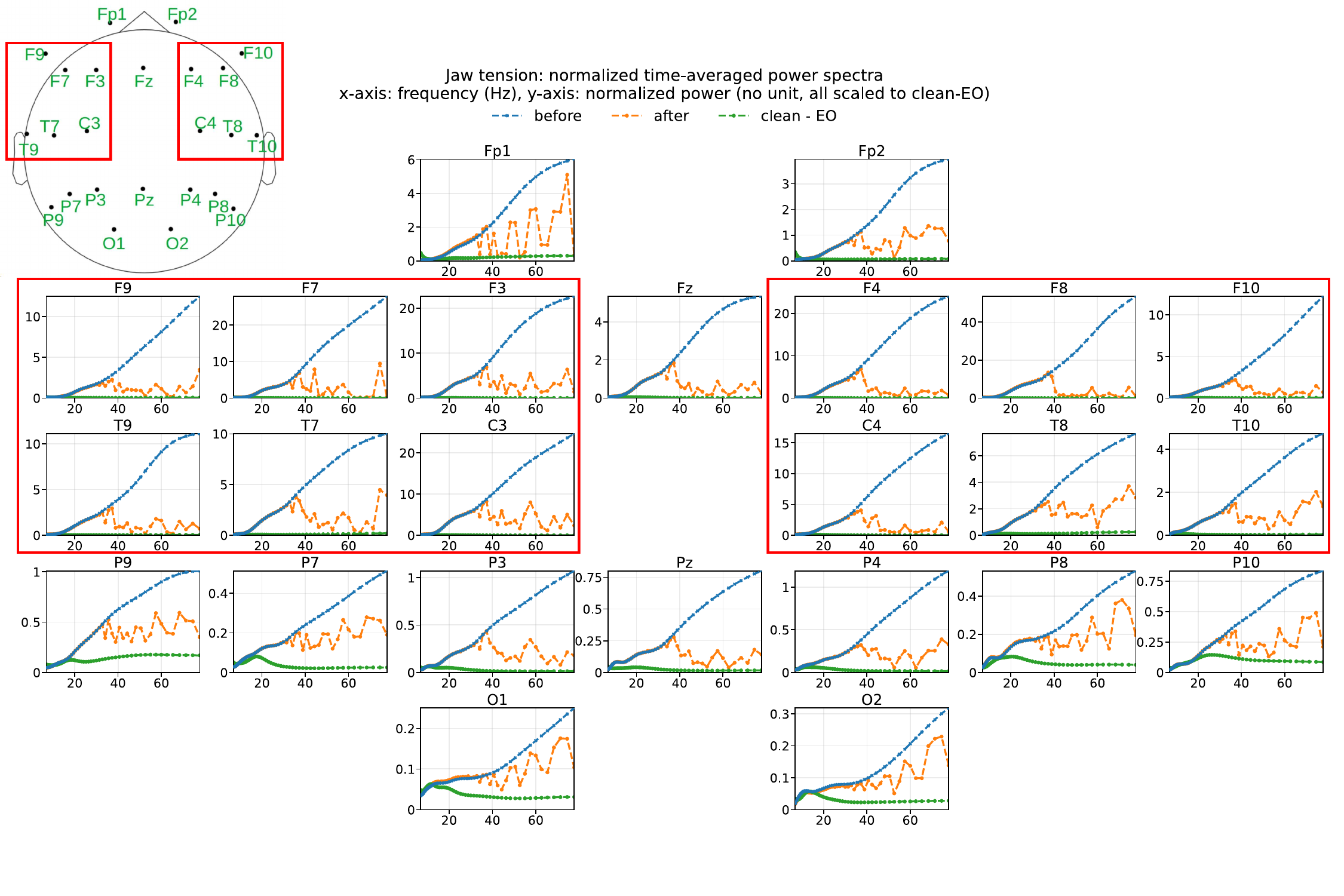}
        \caption{Representative jaw-tension artifact example showing normalized time-averaged power spectra before and after removal of artifact-like intra-components. Each panel corresponds to one EEG channel (refer to the topomap upper right) and shows the artifact-contaminated epoch before cleaning, the reconstructed epoch after cleaning, and the EO reference spectrum. Importantly, complete overlap between the after-cleaning curve and the clean reference curve is not expected. The y-axis is scaled independently for each channel to emphasize channel-specific spectral changes. Red rectangles highlight frontal, fronto-temporal, and central channels in which jaw tension produced a pronounced broadband increase in power, particularly at higher frequencies. These channels also show the largest y-axis ranges.  After intra-component removal, the power spectra were strongly attenuated across these channels and shifted substantially closer to the EO reference, with smaller reductions also visible over other channels.}
        \label{fig:area_diff_improve_jaw}
        \end{figure}

\begin{figure}[t!]
        \includegraphics[width=\linewidth, trim=0 20mm 0 0mm, clip ]{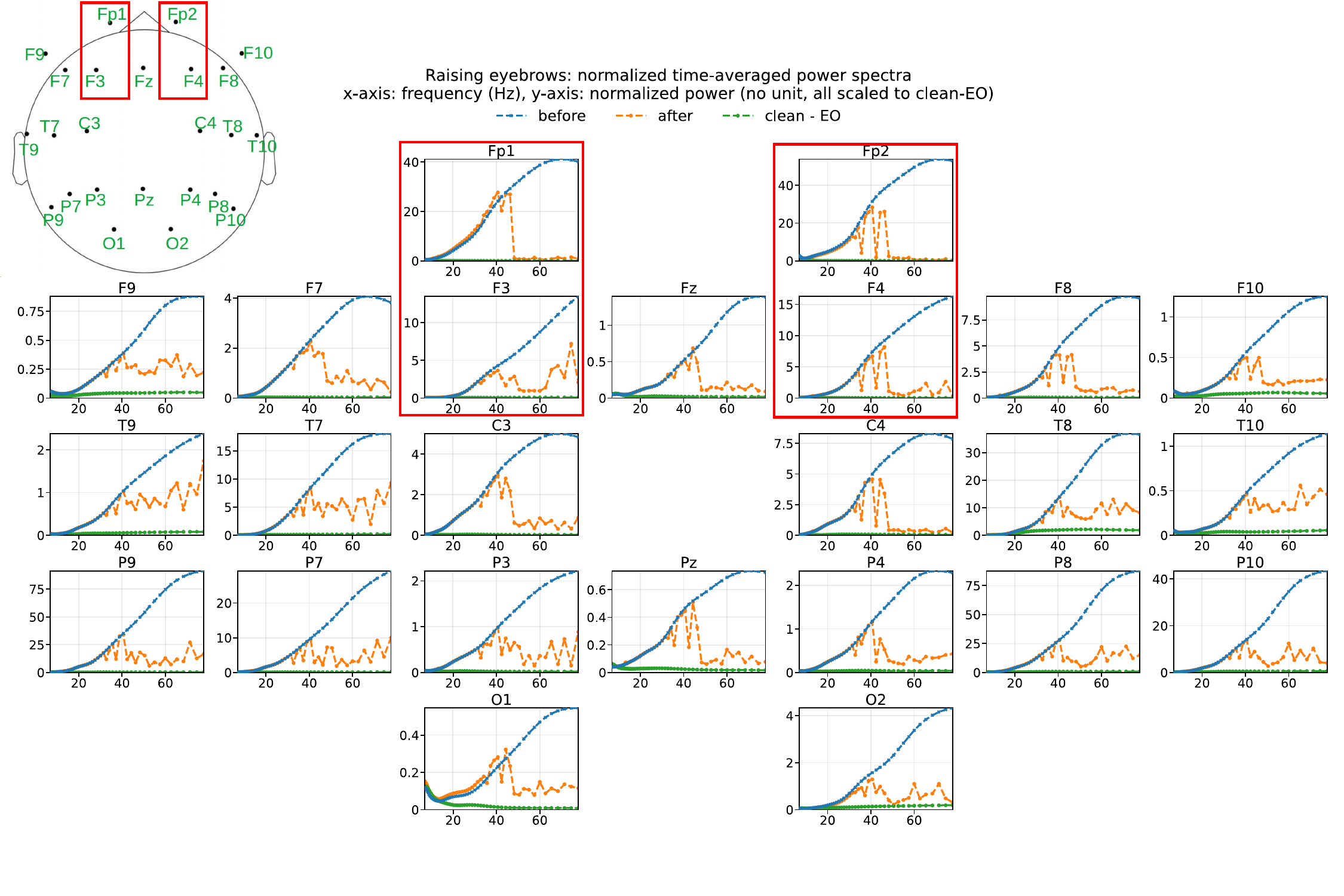}
        \caption{Representative raising-eyebrows artifact example showing normalized time-averaged power spectra before and after removal of artifact-like intra-components. Each panel corresponds to one EEG channel (refer to the topomap upper right) and shows the artifact-contaminated epoch before cleaning, the reconstructed epoch after cleaning, and the EO reference spectrum. Importantly, complete overlap between the after-cleaning curve and the clean reference curve is not necessarily expected. The y-axis is scaled independently for each channel to emphasize channel-specific spectral changes. The red rectangle highlights frontal channels in which the raising-eyebrows artifact produced elevated broadband power. These channels also show the largest y-axis ranges. 
        After intra-component removal, the spectra were attenuated across most channels, with the strongest visible reduction in the highlighted frontal channels and smaller reductions in non-frontal channels. }
        \label{fig:area_diff_improve_raising}
        \end{figure}

        In addition to the channel-wise summary, two representative examples are shown in \cref{fig:area_diff_improve_jaw,fig:area_diff_improve_raising}. Each subplot shows the normalized time-averaged power spectrum for one EEG channel before cleaning, after intra-component removal, and for the EO reference. Importantly, complete overlap between the after-cleaning curve and the reference is not expected, because artifact epochs were recorded during task performance whereas the EO reference represents rest; residual differences may therefore reflect both remaining artifact and task-related brain-state differences.
        For the jaw tension example, the contaminated spectrum shows a pronounced broadband increase in power, especially at higher frequencies and over the frontal, fronto-temporal, and central channels highlighted by the red rectangles; these channels also show the largest y-axis ranges. After intra-component removal, the spectra in these channels were strongly attenuated and shifted substantially closer to the EO reference, consistent with the large positive area-improvement values.
        In the raising-eyebrows example, elevated broadband power was most visible in frontal channels and was reduced after cleaning, with smaller reductions over non-frontal channels.
        These examples illustrate that the score-guided attenuation suppresses high-power spectral contributions associated with artifact-contaminated epochs while retaining the broad channel-wise spectral profile, with the strongest effect for jaw tension and a more moderate effect for raising-eyebrows artifacts.

\section{Discussion and Prospects for Future Research}
    This work suggests that weak epoch-level supervision can induce useful intra-component artifact-likelihood scores. Suppressing high-scoring intra-components produced measurable spectral attenuation, consistent with the MIL assumption that artifact-contaminated epochs contain a subset of intra-components carrying artifact-related evidence, while others may retain useful EEG information. The effect was strongest for jaw tension, where attenuation reduced spectral deviation from the same-subject EO reference in channels consistent with jaw-related muscle activity. Raising-eyebrows artifacts showed more moderate improvement, whereas frowning showed little improvement, suggesting that separability achieved by the present representation is artifact-dependent and may require artifact-specific or multi-class modeling.

    Several limitations should be considered when interpreting these results. First, the weak labels are defined by experimental condition, so the model may learn signal patterns associated with instructed facial movements in addition to EMG contamination itself. Second, no intra-component-level ground truth is available. The same-subject EO reference is therefore only a practical comparison target, not a true clean counterpart of the artifact epoch, and differences between reconstructed spectra and the reference may reflect residual artifact, task-versus-rest brain-state differences, or neural signal attenuated together with artifact-related activity. Synthetic ground truth could in principle provide direct validation of intra-component artifact scores, but this is currently limited by the absence of an established method for generating validated and realistic multichannel EEG–EMG mixtures. Any such simulation would need to prescribe the source structures and spatial distributions of EMG contamination that the present framework seeks to infer from real recordings. Consequently, synthetic ground-truth validation cannot currently provide a reliable basis for evaluating performance under realistic recording conditions. Evaluation through downstream tasks, such as BCI classification, may therefore offer an additional indirect assessment of whether artifact attenuation improves signal utility while preserving task-relevant neural information.

    Because no matching benchmark or evaluation protocol is currently available, and because the present study does not include direct matched comparisons with existing artifact-removal pipelines under a common dataset and evaluation setup, the present results do not demonstrate superiority over existing artifact-removal methods such as automated ICA classifiers. Rather, the contribution is complementary: the proposed framework examines whether weak epoch-level labels can be converted into fine-grained intra-component scores within a frequency-resolved representation. Future work will include a matched component-level MIL control in which conventional components are treated as instances, allowing the specific contribution of the frequency-resolved intra-component representation to be assessed. Additional experiments will examine the sensitivity of the results to ILRMA and architecture parameters, perform feature-ablation analyses, and test generalizability across datasets, artifact types, mental states, and hardware configurations. Direct comparisons with representative artifact-removal approaches should assess downstream task performance, frequency-band-specific spectral preservation, and computational efficiency. 

\section{Conclusion}

    This work addresses two key challenges in EMG artifact removal in EEG data where (1) neural and muscular activity may remain mixed after blind source separation and (2) reliable (intra-)component-level annotations and realistic simulation-based ground truth are unavailable. We propose a weakly supervised framework that combines a frequency-aware high-dimensional representation with MIL. While the representation creates a space in which mixed neural and muscular activity becomes more separable, the MIL formulation treats each epoch as a bag and each frequency-resolved intra-component as an instance, enabling the model to learn intra-component artifact-likelihood scores from epoch-level labels alone, without requiring finer-grained ground truth.
    The experiments suggest that the trained classifier assigns high artifact-likelihood scores to intra-components whose attenuation reduces spectral deviations in the reconstructed time-frequency representation from the same-subject EO reference. The clearest improvements were observed for jaw tension, with more moderate effects for raising-eyebrows artifacts and limited effects for frowning. 
    
    Overall, the proposed approach provides a proof-of-concept route toward fine-grained EEG artifact attenuation without requiring (intra-)component-level ground truth. By integrating temporal, spatial, and frequency-aware representations into a weak-supervision pipeline, the framework may also be useful for broader neural signal analysis problems where fine-grained annotations are scarce, such as biomarker discovery.

%
%
\bibliographystyle{splncs04}
\bibliography{My_Library,other_refs}

\clearpage
\newpage
\appendix

\begin{center}
    \Large\bfseries Appendix
\end{center}
\vspace{1em}

\setcounter{figure}{0}
\setcounter{table}{0}
\counterwithin{figure}{section} 
\counterwithin{table}{section}

\section{Handcrafted Temporal Features} \label{Appendix_temporal_features}
The handcrafted temporal feature vector contains 66 scalar features per intra-component:
45 magnitude-based features (\cref{appendix_tab:magnitude_temporal_features}) and
21 phase-based features (\cref{appendix_tab:phase_temporal_features}). The column
``Dim.'' gives the number of scalar features contributed by each feature group. Small positive numerical stabilizers were added to logarithms and denominators to avoid undefined values for zero amplitudes or near-zero normalization constants. In the formulas below they are denoted generically by \(\epsilon\); in the implementation, they were fixed to \(10^{-8}\) for logarithms, z-scoring, and windowed max-to-median ratios, and to \(10^{-12}\) for autocorrelation, top-energy normalization, and phase-weight normalization.

Let \(Y^*_{fstc}=A_{fcs}Y_{fst}\) denote the high-dimensional intra-component
tensor described in the main text. For a fixed frequency bin \(f\) and
component (ILRMA source) \(s\), one intra-component corresponds to the frequency-component pair
\((f,s)\). Its temporal features were computed from the associated complex source
activation \(Y_{fst}\), which is shared across channels before multiplication by
the channel-specific mixing coefficients \(A_{fcs}\). For notational simplicity,
we write this one-dimensional time series as \(y(t)=Y_{fst}\).
The magnitude representation is computed as \(m(t)=\log(|y(t)|+\epsilon)\), linearly resampled to \(T=256\) samples, and z-scored
to obtain \(x(t)=(m(t)-\mu_m)/(\sigma_m+\epsilon)\). Here,
\(\mu_m\) and \(\sigma_m\) are the mean and standard deviation of the resampled
log-magnitude \(m(t)\). The burst threshold is
\(\tau=\mathrm{median}(x)+1.5\,\mathrm{std}(x)\). The indicator \(\mathbf{1}[\cdot]\) equals one when its condition is true and zero otherwise. The interval \([u_j,z_j]\) denotes the \(j\)-th
contiguous above-threshold segment. The operator \(\mathrm{Top}_{10\%}\) denotes the indices of the largest \(\mathrm{round}(0.1T)\) values of \(\exp(x(t))\), corresponding to 26 samples
when \(T=256\). The smoothed magnitude \(\tilde{x}(t)\) is obtained by convolving \(x(t)\) with a length-\(L\) uniform moving-average kernel \(g\), where \(g(h)=1/L\) for
\(h=1,\ldots,L\). For lag-based features, \(\ell\in\{0.02,0.05,0.10,0.20\}\) denotes a relative lag and \(d_\ell=\mathrm{round}(\ell(T-1))\) the corresponding sample lag.
The window length is \(L=\max(3,\mathrm{round}(0.03T))\), adjusted to the next odd value; here \(L=9\). Windowed features use \(K\in\{4,8,16\}\) equal temporal windows \(W_r\), where \(r\)
indexes the window. 

The raw phase is \(\theta_{\mathrm{raw}}(t)=\arg(y(t))\), which is linearly resampled
to the normalized length \(T=256\) to obtain \(\theta(t)\). The wrapped phase
increments are defined as
\(\Delta\theta(t)=\arg\{\exp(j[\theta(t+1)-\theta(t)])\}\), where \(j\) is the imaginary
unit and \(\exp(\cdot)\) denotes the exponential function.
For \(W_r=\{a_r,\ldots,b_r\}\), windowed phase increments are defined as
$\Delta\theta_r(t)=\arg\{\exp(j[\theta(t+1)-\theta(t)])\}, t=a_r,\ldots,b_r-1,$
so increments crossing window boundaries are excluded. 

\begin{table}[!tbp]
\caption{Handcrafted magnitude-based temporal features used for each intra-component. The column ``Dim.'' gives the number of scalar features contributed by each feature group.}
\label{appendix_tab:magnitude_temporal_features}
\centering
\setlength{\tabcolsep}{3pt}
\renewcommand{\arraystretch}{1.12}
\begin{tabularx}{\linewidth}{|p{0.20\linewidth}|p{0.30\linewidth}|X|p{0.06\linewidth}|}
\hline
Feature group & Formula & Short explanation & Dim. \\
\hline
Mean magnitude activity &
\(\frac{1}{T}\sum_{t=1}^{T}|x(t)|\) &
Average absolute normalized log-magnitude. &
1 \\
\hline
Burst fraction &
\(\frac{1}{T}\sum_{t=1}^{T}\mathbf{1}[x(t)>\tau]\) &
Fraction of samples above the burst threshold. &
1 \\
\hline
Longest burst duration &
\(\frac{1}{T}\max_j(z_j-u_j+1)\) &
Longest contiguous above-threshold segment, normalized by signal length. &
1 \\
\hline
High-energy concentration &
\(\frac{\sum_{t\in \mathrm{Top}_{10\%}}\exp(x(t))}
{\sum_{t=1}^{T}\exp(x(t))+\epsilon}\) &
Fraction of magnitude energy contained in the largest 10\% of samples. &
1 \\
\hline
Temporal variation &
\(\frac{1}{T-1}\sum_{t=1}^{T-1}|\tilde{x}(t+1)-\tilde{x}(t)|\) &
Mean absolute first difference of the smoothed magnitude signal. &
1 \\
\hline
Magnitude autocorrelation &
\(
\rho_x(\ell)= \newline 
\frac{
\sum_{t=1}^{T-d_\ell}
\left(x(t)-\bar{x}\right)
\left(x(t+d_\ell)-\bar{x}\right)
}{
(T-d_\ell)\left(\mathrm{var}(x)+\epsilon\right)+\epsilon
}, \newline
\ell\in\{0.02,0.05,0.10,0.20\}\) &
Autocorrelation of \(x(t)\) at four relative lags. \(\bar{x}\) is the mean of \(x(t)\) and $\mathrm{var}(x)$ the variance. &
4 \\
\hline
Windowed magnitude mean statistics &
\(\max_r \bar{x}_r,\ \mathrm{std}_r(\bar{x}_r),\ \newline 
\frac{\max_r\bar{x}_r}{\mathrm{median}_r(\bar{x}_r)+\epsilon}\) &
For each \(K\), summarize window-wise means \(\bar{x}_r\). &
9 \\
\hline
Windowed magnitude peak statistics &
\(\max_r \alpha_r,\ \mathrm{std}_r(\alpha_r),\ \newline 
\frac{\max_r \alpha_r}{\mathrm{median}_r(\alpha_r)+\epsilon}\) &
For each \(K\), summarize window-wise peaks \(\alpha_r=\max_{t\in W_r}x(t)\). &
9 \\
\hline
Windowed magnitude variance statistics &
\(\max_r \sigma_r^2,\ \mathrm{std}_r(\sigma_r^2),\ \newline 
\frac{\max_r\sigma_r^2}{\mathrm{median}_r(\sigma_r^2)+\epsilon}\) &
For each \(K\), summarize window-wise variances \(\sigma_r^2\). &
9 \\
\hline
Windowed burst statistics &
\(\max_r \beta_r,\ \mathrm{std}_r(\beta_r),\ \newline 
\frac{\max_r \beta_r}{\mathrm{median}_r(\beta_r)+\epsilon}\) &
For each \(K\), summarize window-wise burst fractions \(\beta_r\). &
9 \\
\hline
Total & & & 45 \\
\hline
\end{tabularx}
\end{table}

\begin{table}[!tbp]
\caption{Handcrafted phase-based temporal features used for each intra-component. The column ``Dim.'' gives the number of scalar features contributed by each feature group.}
\label{appendix_tab:phase_temporal_features}
\centering
\setlength{\tabcolsep}{3pt}
\renewcommand{\arraystretch}{1.12}
\begin{tabularx}{\linewidth}{|p{0.20\linewidth}|p{0.32\linewidth}|X|p{0.08\linewidth}|}
\hline
Feature group & Formula & Short explanation & Dim. \\
\hline
Circular concentration &
\(\left|\frac{1}{T}\sum_{t=1}^{T}\exp(j\theta(t))\right|\) &
Circular resultant length of the phase sequence. &
1 \\
\hline
Phase-increment concentration &
\(\left|\frac{1}{T-1}\sum_{t=1}^{T-1}\exp(j\Delta\theta(t))\right|\) &
Circular resultant length of phase increments. &
1 \\
\hline
Phase increment variability &
\(\mathrm{std}_t(\Delta\theta(t))\) &
Standard deviation of wrapped phase increments. &
1 \\
\hline
Mean absolute phase increment &
\(\frac{1}{T-1}\sum_{t=1}^{T-1}|\Delta\theta(t)|\) &
Average absolute phase change between adjacent samples. &
1 \\
\hline
Amplitude-weighted phase concentration &
\(\left|\sum_{t=1}^{T} w_t \exp(j\theta(t))\right|,
w_t= \newline
\frac{\exp(x(t)-x_{\max})}
{\sum_{v=1}^{T}\exp(x(v)-x_{\max})+\epsilon},
\newline
x_{\max}=\max_{1\le r\le T}x(r)\) &
Circular resultant length of phase, weighted by amplitude energy. &
1 \\
\hline
Phase similarity &
\(\frac{1}{T-d_\ell}\sum_{t=1}^{T-d_\ell}
\cos(\theta(t+d_\ell)-\theta(t)) , \newline \
\ell\in\{0.02,0.05,0.10,0.20\}\) &
Mean phase similarity at four relative lags. &
4 \\
\hline
Windowed absolute phase-increment statistics &
\(\max_r q_r,\ \mathrm{std}_r(q_r)\) &
For each \(K\), summarize \(q_r =
\frac{1}{|W_r|-1}
\sum_{t=a_r}^{b_r-1}
|\Delta\theta_r(t)|\) &
6 \\
\hline
Windowed phase-increment concentration statistics &
\(\max_r c_r,\ \mathrm{std}_r(c_r)\) &
For each \(K\), summarize
\(c_r =
\left|
\frac{1}{|W_r|-1}
\sum_{t=a_r}^{b_r-1}
\exp(j\Delta\theta_r(t))
\right|\) &
6 \\
\hline
Total & & & 21 \\
\hline
\end{tabularx}
\end{table}

\section{Detailed Architecture of the Intra-Component Classifier} \label{Appendix_architecture}
The architecture of the intra-component classifier is summarized in \cref{appendix_tab:mil_architecture} and illustrated in \cref{appendix_fig:arch}. Each intra-component is represented by three inputs: a spatial topomap, a handcrafted temporal feature vector, and a normalized frequency coordinate. These inputs are encoded by shared spatial, temporal, and fusion modules, followed by a component-wise frequency-context encoder. The final sigmoid score head assigns one artifact-likelihood score to each intra-component.

\begin{table}[!tbp]
\caption{Detailed architecture of the intra-component classifier. All convolutional and linear layers use learned biases; no batch normalization is used, with LayerNorm applied only to the 66-D temporal feature input. Here, $F$ denotes the number of frequency bins, $S$ denotes the number of components (ILRMA sources), and $N=F\times S$ is the number of intra-components in one MIL bag, corresponding to one EEG epoch.}
\label{appendix_tab:mil_architecture}
\centering
\setlength{\tabcolsep}{3pt}
\renewcommand{\arraystretch}{1.12}
\begin{tabularx}{\linewidth}{|p{0.20\linewidth}|p{0.22\linewidth}|p{0.16\linewidth}|p{0.16\linewidth}|X|}
\hline
Module & Layer & Kernel/Stride\newline/Padding & Activation\newline /Dropout & Output \\
\hline
Input & Spatial \newline topomaps $T$ & -- & -- & $N{\times}3{\times}32{\times}32$ \\
Spatial & Conv2D $3{\to}16$ & $3{\times}3 / 1 / 1$ & GELU & $N{\times}16{\times}32{\times}32$ \\
Spatial & Conv2D $16{\to}32$ & $3{\times}3 / 2 / 1$ & GELU & $N{\times}32{\times}16{\times}16$ \\
Spatial & Conv2D $32{\to}64$ & $3{\times}3 / 2 / 1$ & GELU & $N{\times}64{\times}8{\times}8$ \\
Spatial & Adaptive avg. \newline pool + flatten & $1{\times}1$ & Dropout 0.1 & $N{\times}64$ \\
Spatial & Linear $64{\to}64$ & -- & GELU  & $N{\times}64$ \\
\hline
Input & Temporal \newline vectors $V$ & -- & -- & $N{\times}66$ \\
Temporal & LayerNorm + \newline linear $66{\to}64$ & -- & GELU,\newline Dropout 0.1 & $N{\times}64$ \\
Temporal & Linear $64{\to}64$ & -- & GELU & $N{\times}64$ \\
\hline
Input & Spatial + \newline temporal output + frequency \newline coordinates $f$ & -- & -- & $N{\times}64$, $N{\times}64$, $N{\times}1$ \\
Concatenate & $64+64+1 {\to} 129$ & -- & -- & $N{\times}129$\\
\hline
Fusion & Linear $129{\to}64$ & -- & GELU,\newline Dropout 0.1 & $N{\times}64$ \\
Embedding & Linear $64{\to}64$ & -- & GELU & $H \in \mathbb{R}^{N{\times}64}$ \\
\hline
Frequency\newline context & Reshape per \newline component\newline(ILRMA source) \newline $[N,64] \to [S,64,F]$ & -- & -- & $S{\times}64{\times}F$ \\
Frequency\newline context & Conv1D $64{\to}64$ & $5 / 1 / 2$ & GELU,\newline Dropout 0.1 & $S{\times}64{\times}F$ \\
Frequency\newline context & Conv1D $64{\to}64$ & $5 / 1 / 2$ & GELU & $S{\times}64{\times}F$ \\
Frequency\newline context & Reshape back + \newline residual add & -- & -- & $\tilde{H} \in \mathbb{R}^{N{\times}64}$ \\
\hline
Score head & Linear $64{\to}1$ & -- & Sigmoid & $p \in [0,1]^{N{\times}1}$ \\
\hline
\end{tabularx}

\vspace{0.25em}
\footnotesize
\end{table}

\begin{figure}[!tbp]
    \centering
    \includegraphics[width=\linewidth, trim=13mm 115mm 12mm 0mm, clip]{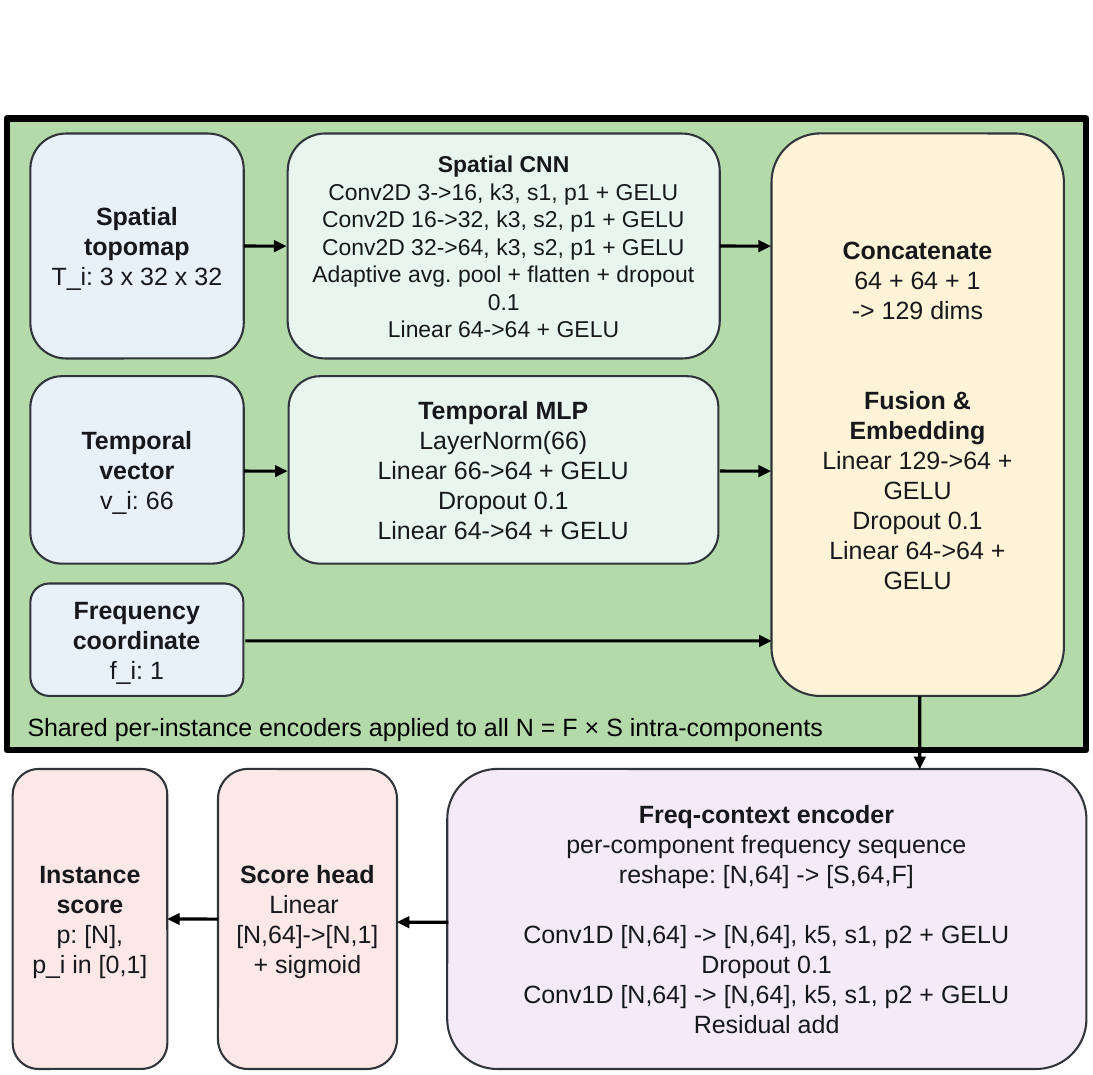}
    \caption{Detailed architecture of the intra-component classifier. Each intra-component has three inputs: a 3-channel spatial topomap \(T_i \in \mathbb{R}^{3 \times 32 \times 32}\), a 66-dimensional handcrafted temporal feature vector \(v_i \in \mathbb{R}^{66}\), and a normalized scalar frequency coordinate \(f_i \in \mathbb{R}\). Kernel size, stride, and padding are denoted by $k$, $s$, and $p$, respectively. The spatial CNN, temporal MLP, and fusion modules are shared per-instance encoders applied to all $N=F\times S$ intra-components, where \(F\) is the number of frequency bins and \(S\) is the number of ILRMA sources, as indicated by the green block. The frequency-context encoder then gathers all embeddings in the bag (one EEG epoch), reshapes them to $[S,64,F]$, and applies 1-D convolutions along the frequency axis for each component (ILRMA source). Each final score $p_i$ is therefore an intra-component artifact-likelihood score; one EEG epoch produces $N$ such scores.}
    \label{appendix_fig:arch}
\end{figure}

\section{Detailed Descriptive Statistics of Relative Area-Difference Improvement} \label{Appendix_stats}
 
This subsection reports channel-wise descriptive statistics for the relative area-difference improvement \(\Delta d_c\) after score-guided removal of artifact-like intra-components, grouped by artifact type: jaw tension (JT), raising eyebrows (RE), and frowning (Fr), with 24 epochs per type (\cref{appendix_tab:area_diff_improvement_rel_stats_kb_a,appendix_tab:area_diff_improvement_rel_stats_sh_a,appendix_tab:area_diff_improvement_rel_stats_sr_a}). For each channel, improvement was computed by comparing the spectral area difference from the same-subject EO reference before and after removal. Note that the EO reference serves as a practical comparison target, rather than a true ground truth. Differences between the cleaned spectrum and the clean reference may reflect both residual artifact and genuine differences in brain state. Positive values indicate reduced deviation from the EO reference, whereas negative values indicate increased deviation. The columns ``p5'', ``p25'', ``p75'', and ``p95'' denote the 5th, 25th, 75th, and 95th percentiles, respectively, and the median column reports the 50th percentile across evaluated epochs.

\begin{table}[!tbp]
\caption{Channel-wise relative area-difference improvement after score-guided removal of artifact-like intra-components for jaw tension (JT).}
\label{appendix_tab:area_diff_improvement_rel_stats_kb_a}
\begin{center}
\begin{tabular}{|l|l|c|c|c|c|c|}
\hline
Artifact class & Channel & $p5$ & $p25$ & Median & $p75$ & $p95$ \\
\hline\hline
JT & Fp1 & -6.13 & 0.04 & 5.09 & 13.39 & 68.34 \\
JT & Fp2 & 0.31 & 1.70 & 7.38 & 16.72 & 67.29 \\
JT & F9 & 0.91 & 8.81 & 84.77 & 329.51 & 717.42 \\
JT & F7 & 1.54 & 8.93 & 111.53 & 291.32 & 781.37 \\
JT & F3 & 0.98 & 13.96 & 86.69 & 222.52 & 648.78 \\
JT & Fz & 1.79 & 9.72 & 25.21 & 91.12 & 156.08 \\
JT & F4 & 3.67 & 22.65 & 120.05 & 184.52 & 422.13 \\
JT & F8 & 4.56 & 29.41 & 144.94 & 520.61 & 707.97 \\
JT & F10 & 5.94 & 26.64 & 141.23 & 305.05 & 763.93 \\
JT & T9 & 0.26 & 3.95 & 43.53 & 150.15 & 597.23 \\
JT & T7 & 0.12 & 4.35 & 18.54 & 65.94 & 259.32 \\
JT & C3 & 1.49 & 11.85 & 95.97 & 278.09 & 625.51 \\
JT & C4 & 5.40 & 21.31 & 124.30 & 289.53 & 604.46 \\
JT & T8 & 0.45 & 2.95 & 14.68 & 86.38 & 859.18 \\
JT & T10 & 1.18 & 6.72 & 42.83 & 105.76 & 851.62 \\
JT & P9 & 0.06 & 0.46 & 1.52 & 3.27 & 35.94 \\
JT & P7 & 0.08 & 0.71 & 2.53 & 31.08 & 157.01 \\
JT & P3 & 0.68 & 2.49 & 11.75 & 50.13 & 169.01 \\
JT & Pz & 0.55 & 1.99 & 11.10 & 40.10 & 91.82 \\
JT & P4 & 1.02 & 5.03 & 16.89 & 70.17 & 170.95 \\
JT & P8 & 0.09 & 0.40 & 4.22 & 24.66 & 138.19 \\
JT & P10 & 0.02 & 0.10 & 0.70 & 4.21 & 22.14 \\
JT & O1 & 0.07 & 0.54 & 2.47 & 8.52 & 30.47 \\
JT & O2 & 0.20 & 0.34 & 1.45 & 10.16 & 21.99 \\
\hline
\end{tabular}
\end{center}
\end{table}

\begin{table}[!tbp]
\caption{Channel-wise relative area-difference improvement after score-guided removal of artifact-like intra-components for raising eyebrows (RE).}
\label{appendix_tab:area_diff_improvement_rel_stats_sh_a}
\begin{center}
\begin{tabular}{|l|l|c|c|c|c|c|}
\hline
Artifact class & Channel & $p5$ & $p25$ & Median & $p75$ & $p95$ \\
\hline\hline
RE & Fp1 & -35.51 & -6.57 & -0.47 & 104.93 & 1326.41 \\
RE & Fp2 & 0.42 & 3.35 & 17.97 & 145.74 & 656.11 \\
RE & F9 & -0.07 & 0.08 & 0.30 & 3.01 & 17.68 \\
RE & F7 & 0.03 & 0.24 & 2.27 & 14.30 & 216.86 \\
RE & F3 & 0.01 & 2.43 & 9.86 & 98.71 & 497.83 \\
RE & Fz & 0.05 & 1.02 & 3.03 & 43.88 & 91.38 \\
RE & F4 & 0.01 & 1.77 & 10.03 & 116.20 & 234.43 \\
RE & F8 & 0.04 & 0.25 & 3.33 & 13.14 & 77.42 \\
RE & F10 & 0.00 & 0.09 & 0.66 & 4.33 & 26.20 \\
RE & T9 & 0.01 & 0.07 & 0.31 & 0.90 & 8.98 \\
RE & T7 & -0.00 & 0.06 & 0.52 & 1.33 & 4.85 \\
RE & C3 & 0.03 & 0.76 & 1.89 & 10.23 & 70.97 \\
RE & C4 & 0.05 & 0.86 & 1.72 & 9.96 & 61.83 \\
RE & T8 & 0.00 & 0.23 & 1.06 & 5.43 & 17.55 \\
RE & T10 & 0.00 & 0.06 & 0.31 & 1.14 & 13.17 \\
RE & P9 & 0.00 & 0.12 & 0.32 & 1.75 & 32.23 \\
RE & P7 & -0.01 & 0.08 & 0.55 & 3.82 & 46.88 \\
RE & P3 & 0.00 & 0.22 & 0.70 & 3.44 & 29.19 \\
RE & Pz & 0.01 & 0.19 & 0.83 & 4.73 & 15.81 \\
RE & P4 & 0.01 & 0.22 & 0.76 & 4.13 & 23.75 \\
RE & P8 & 0.01 & 0.16 & 0.90 & 2.28 & 21.40 \\
RE & P10 & 0.00 & 0.10 & 0.46 & 1.23 & 29.13 \\
RE & O1 & -0.11 & 0.08 & 0.67 & 1.53 & 5.65 \\
RE & O2 & 0.00 & 0.18 & 0.54 & 1.78 & 3.67 \\
\hline
\end{tabular}
\end{center}
\end{table}

\begin{table}[!tbp]
\caption{Channel-wise relative area-difference improvement after score-guided removal of artifact-like intra-components for frowning (Fr).}
\label{appendix_tab:area_diff_improvement_rel_stats_sr_a}
\begin{center}
\begin{tabular}{|l|l|c|c|c|c|c|}
\hline
Artifact class & Channel & $p5$ & $p25$ & Median & $p75$ & $p95$ \\
\hline\hline
Fr & Fp1 & -8.68 & -1.75 & -0.71 & -0.13 & 11.92 \\
Fr & Fp2 & 0.00 & 0.03 & 0.24 & 0.74 & 4.29 \\
Fr & F9 & -0.03 & -0.00 & 0.02 & 0.06 & 0.32 \\
Fr & F7 & -0.00 & 0.00 & 0.02 & 0.06 & 0.40 \\
Fr & F3 & -0.01 & 0.00 & 0.06 & 0.14 & 0.72 \\
Fr & Fz & -0.01 & 0.00 & 0.02 & 0.04 & 2.06 \\
Fr & F4 & -0.00 & 0.00 & 0.02 & 0.07 & 2.28 \\
Fr & F8 & -0.04 & 0.00 & 0.04 & 0.14 & 0.36 \\
Fr & F10 & -0.00 & 0.01 & 0.03 & 0.07 & 0.63 \\
Fr & T9 & -0.03 & -0.01 & 0.00 & 0.04 & 0.27 \\
Fr & T7 & -0.08 & -0.00 & 0.00 & 0.04 & 0.48 \\
Fr & C3 & -0.01 & 0.00 & 0.02 & 0.06 & 0.26 \\
Fr & C4 & -0.02 & 0.00 & 0.01 & 0.06 & 0.41 \\
Fr & T8 & -0.06 & -0.01 & 0.00 & 0.02 & 0.66 \\
Fr & T10 & -0.05 & -0.01 & 0.01 & 0.07 & 0.39 \\
Fr & P9 & -0.03 & -0.00 & 0.00 & 0.04 & 0.21 \\
Fr & P7 & -0.02 & 0.00 & 0.02 & 0.06 & 0.44 \\
Fr & P3 & -0.02 & 0.00 & 0.00 & 0.03 & 0.50 \\
Fr & Pz & -0.01 & 0.00 & 0.02 & 0.04 & 0.42 \\
Fr & P4 & -0.07 & 0.00 & 0.02 & 0.03 & 0.35 \\
Fr & P8 & -0.04 & 0.00 & 0.00 & 0.04 & 0.20 \\
Fr & P10 & -0.79 & -0.02 & 0.00 & 0.02 & 0.37 \\
Fr & O1 & -0.06 & -0.01 & 0.00 & 0.03 & 0.14 \\
Fr & O2 & -0.23 & -0.02 & -0.00 & 0.02 & 0.45 \\
\hline
\end{tabular}
\end{center}
\end{table}

\end{document}